\documentclass[lettersize,journal]{IEEEtran}
\usepackage{amsmath,amsfonts}
\usepackage{algorithmic}
\usepackage{algorithm}
\usepackage{array}
\usepackage[caption=false,font=normalsize,labelfont=sf,textfont=sf]{subfig}
\usepackage{textcomp}
\usepackage{stfloats}
\usepackage{url}
\usepackage{verbatim}
\usepackage{graphicx}
\usepackage{cite}
\usepackage{pifont}
\newcommand{\cmark}{\ding{51}}%
\usepackage{tabularx} % 导言区添加

\usepackage{multirow}
\usepackage{cuted}%%\stripsep-3pt
\usepackage{booktabs}
\usepackage{dutchcal} % \mathcal: also small letters. \mathbcal = bold ones. 

\usepackage{hyperref}       % hyperlinks
\usepackage{mathrsfs}
\usepackage[dvipsnames]{xcolor}

\begin{document}

\title{ARC-Fi: Exploiting Antenna Spatial Diversity for Label-Efficient Domain Generalization in Wi-Fi Sensing}

\author{Ke Xu, Zhiyong Zheng, Hongyuan Zhu,~\IEEEmembership{Member,~IEEE}, Lei Wang, ~\IEEEmembership{Member,~IEEE}, Jiangtao Wang*,~\IEEEmembership{Member,~IEEE}

\thanks{Ke Xu is with the Suzhou Institute for Advanced Research, University of Science and Technology of China, Suzhou, China, and also with the Suzhou Big Data and AI Research and Engineering Center, Suzhou, China.  (e-mails: nick\_xuke@ustc.edu.cn).}%
\thanks{Zhiyong Zheng is with the School of Artificial Intelligence and Data Science, University of Science and Technology of China, Hefei, China (e-mail: zhiyongzheng@mail.ustc.edu.cn).}%
\thanks{Hongyuan Zhu is with the Institute for Infocomm Research (I\textsuperscript{2}R), A*STAR, Singapore (e-mail: zhuh@i2r.a-star.edu.sg).}%
\thanks{Lei Wang is with the School of Computer Science and Technology, Soochow University, Suzhou, China (e-mail: wanglei@suda.edu.cn).}%
\thanks{Manuscript received October 19, 2025. (Corresponding author: Jiangtao Wang.)}%
}

% The paper headers
\markboth{Journal of \LaTeX\ Class Files,~Vol.~0, No.~0, August~2026}%
%{Xu \MakeLowercase{\textit{et al.}}: \textbf{ARC-Fi}: A Unified Framework For Cross-Domain Wi-Fi-based Gesture Recognition}
{Shell \MakeLowercase{\textit{et al.}}: A Sample Article Using IEEEtran.cls for IEEE Journals}

%\IEEEpubid{0000--0000/00\$00.00~\copyright~2021 IEEE}
% Remember, if you use this you must call \IEEEpubidadjcol in the second
% column for its text to clear the IEEEpubid mark.

\maketitle

\begin{abstract}
Wi-Fi-based gesture recognition sensing systems are severely hindered by domain shifts in real-world environments. Existing Unsupervised Domain Adaptation (UDA) and Domain Generalization (DG) methods rely on either inaccessible target data or prohibitively expensive massive labeled source datasets. Since collecting unlabeled Channel State Information (CSI) is feasible but manual labeling is constrained, we address the practical Semi-Supervised Domain Generalization (SSDG) problem by proposing \textbf{ARC-Fi}. While contrastive learning is ideal for leveraging abundant unlabeled data, applying it to raw CSI faces two coupled bottlenecks. First, models are highly susceptible to static environmental "shortcuts," trivially memorizing room signatures rather than gesture dynamics. Second, residual dynamic domain shifts clash with conventional contrastive algorithms, which inherently destroy physical signal semantics during data augmentation and blindly repel same-class instances across domains. To resolve these, \textbf{ARC-Fi} introduces a tightly coupled representation-optimization architecture. To defeat static shortcuts, a physics-guided disentanglement pipeline explicitly strips away macroscopic environmental confounders. To tackle the dynamic algorithmic hurdles, we propose the Antenna Response Consistency (ARC) module at the representation level, which exploits multi-antenna spatial diversity to generate robust, semantics-preserving augmented views. At the optimization level, a tailored Semi-Supervised Contrastive Objective leverages scarce labels and reliable pseudo-labels to correctly align cross-domain features. Extensive experiments on popular public datasets demonstrate that \textbf{ARC-Fi} establishes a new state-of-the-art across SSDG, DG, and UDA scenarios. Code is available at: https://github.com/KaoruMiyazono/UniCrossFi.
\end{abstract}

\begin{IEEEkeywords}
Semi-Supervised Domain Generalization, gesture recognition, Contrastive Learning, Wireless Sensing.
\end{IEEEkeywords}

\section{Introduction}
\label{sec:intro}

Over the past decade, Wi-Fi-based gesture recognition has matured from concept to commercial reality, repurposing Wi-Fi from a communication medium into a powerful sensing modality. Fueled by inherent advantages—such as cost-effectiveness, privacy preservation, and its device-free nature—this technology now enables a wide range of applications, including intuitive control in smart homes, non-intrusive patient monitoring, and seamless human-computer interaction.
More recently, research has shifted from traditional model-driven methods \cite{pu2013whole,abdelnasser2015wigest} to deep learning approaches\cite{jiao2024wisda,liu2023generalizing}.
However, deep neural networks rely heavily on the assumption of independent and identically distributed (i.i.d.) data. 
In real-world applications, when the target test data originate from a distribution different from that of the training data (e.g., a new room or a new user), model performance often degrades significantly.
This phenomenon, known as domain shift \cite{panigrahi2020survey}, is a prevalent challenge in Wi-Fi sensing \cite{zhang2022practical}.

To combat domain shift, existing studies predominantly focus on two settings: Unsupervised Domain Adaptation (UDA) and Domain Generalization (DG). UDA algorithms \cite{jiao2024wisda} attempt to align feature distributions but strictly require access to unlabeled target-domain data during training, which is often infeasible for off-the-shelf deployments in unforeseen environments. DG algorithms \cite{liu2023wisr,zhang2021wifi,wang2024airfi}, on the other hand, eliminate the need for target data by learning domain-invariant representations across multiple source domains. However, the success of DG relies heavily on the availability of \textit{abundant labeled data} from the source domains \cite{gulrajanisearch}. 
In the context of Wi-Fi sensing, fulfilling this requirement is exceptionally challenging. Unlike visual images, raw Channel State Information (CSI) signals are not human-interpretable \cite{li2022unsupervised}. Annotating CSI data cannot be done post-hoc by human annotators; it requires concurrent video recording, specialized synchronization equipment, and labor-intensive alignment during the collection phase. Consequently, obtaining large-scale, diverse, and fully labeled source-domain datasets is prohibitively expensive.

% \textit{Motivation: Bridging the gap with Semi-Supervised Domain Generalization (SSDG).} 
In real-world Wi-Fi deployments, a far more practical scenario emerges: we can easily collect vast amounts of raw, unlabeled CSI data across various source environments (e.g., daily background activities in different smart homes), but we can only afford to manually label a very small subset of it. This practical dilemma gives rise to the SSDG setting. % (as illustrated in Figure~\ref{fig:ProbSet}). 
While SSDG has gained traction in computer vision, it remains largely unexplored in Wi-Fi sensing. Developing a framework specifically tailored to tackle the SSDG challenge is therefore both practically urgent and academically valuable.

To leverage the abundant unlabeled data in the SSDG setting, contrastive learning \cite{chen2020simple,he2020momentum} emerges as a natural choice. However, effectively applying it to CSI data requires navigating two coupled bottlenecks:
\textbf{Challenge 1: The Contrastive Paradigm's Susceptibility to Environmental Shortcuts.} Raw CSI heavily entangles human motion with dominant static multipath. In contrastive learning, the model prioritizes instance discrimination, inevitably exploiting a "static shortcut"—trivially minimizing loss by grouping samples based on energy-heavy room signatures rather than subtle gesture dynamics. This paradigm-specific flaw causes the model to merely memorize specific environments, leading to catastrophic representation collapse under cross-domain scenarios.
\textbf{Challenge 2: Dynamic Domain Shifts and Algorithmic Hurdles.} Even after removing static backgrounds, residual dynamic signals remain entangled with domain-specific factors (e.g., varying user body sizes, locations, and dynamic multipath). Learning domain-invariant semantics under SSDG thus faces two algorithmic hurdles:
1) \textit{The Augmentation Dilemma}: Contrastive learning requires data augmentation to construct positive pairs, but generic augmentations (e.g., time flipping) destroy the physical semantic integrity of CSI, while heuristic signal processing alters core spatio-temporal patterns.
2) \textit{Cross-Domain Misalignment}: Conventional unsupervised objectives indiscriminately repel all negative pairs, blindly pushing apart samples from different domains even when they share the identical gesture semantics.

To this end, we propose \textbf{ARC-Fi}, a robust framework to resolve these challenges.
\textbf{1) For solving challenge 1:} To safeguard the contrastive objective, we introduce a disentanglement pipeline as a necessary prerequisite. By systematically applying phase calibration and static path suppression, we explicitly strip away macroscopic environmental confounders. While these are established signal processing techniques, integrating them as a foundational sanitization step is a crucial insight for cross-domain contrastive learning; it explicitly blocks the trivial static shortcuts and forces the subsequent network to focus strictly on human-induced dynamic multipath.
\textbf{2) Core Innovation for Challenge 2: The ARC Framework.} Building upon this clean foundation, we resolve the augmentation dilemma and cross-domain misalignment inherent in SSDG through a tightly coupled representation-optimization engine. At the \textit{representation level}, we propose our core innovation, the Antenna Response Consistency (ARC) module. It generates semantics-preserving positive views by exploiting the inherent spatial diversity of multi-antenna systems. By synthesizing cross-domain samples via adaptive instance normalization, ARC compels the model to distill pure gesture semantics robust to spatial and user variations. At the \textit{optimization level}, we couple ARC with a tailored Semi-Supervised Contrastive Objective. Specifically designed for the limited-label SSDG setting, it leverages a pseudo-label-guided mechanism to align high-confidence unlabeled samples with scarce supervised source samples, explicitly preventing the blind cross-domain repulsion of same-class instances.

In summary, the main contributions of this work are threefold:
\begin{itemize}
    \item We formally identify and address the practical SSDG problem in Wi-Fi sensing. To the best of our knowledge, \textbf{ARC-Fi} is the first dedicated framework designed to overcome the critical bottleneck of limited labeled CSI data, paving the way for scalable cross-domain deployments in real-world environments.
    \item We propose a tightly coupled representation-optimization architecture. At the representation level, we integrate a foundational physics-guided disentanglement pipeline with our core ARC module to extract robust, semantics-preserving augmentations from multi-antenna spatial diversity. At the optimization level, we couple this with a Unified Semi-Supervised Contrastive Objective that explicitly aligns cross-domain features using scarce labels and reliable pseudo-labels, effectively resolving cross-domain misalignment.
    \item We conduct extensive experiments across diverse real-world settings. The results demonstrate that ARC-Fi consistently achieves state-of-the-art performance in the challenging SSDG setting, alongside superior robustness in conventional DG and UDA scenarios. Furthermore, our detailed ablation studies yield valuable physical insights, confirming that our foundational decoupling explicitly prevents the paradigm-specific environmental shortcut learning inherent to contrastive models.
\end{itemize}

\section{Related Work}
\label{sec:related}
% In this section, we review prior work along two key axes: cross-domain wireless sensing and the application of contrastive learning to CSI data.

\subsection{Cross-Domain Wi-Fi Sensing: From UDA to SSDG}
Exisitng efforts to combat domain shift problem predominantly fall into UDA and DG.
1) UDA attempts to align feature distributions across domains by reducing statistical discrepancies (e.g., MMD) \cite{long2015learning,sun2016return,li2021subdomain,liang2024dcs} or employing adversarial learning to generate domain-invariant features \cite{ganin2016domain,tzeng2017adversarial,jiang2018towards,zou2018joint,kang2021context}. However, UDA algorithms strictly require access to unlabeled target-domain data during training, rendering them impractical for off-the-shelf deployments in unforeseen environments.
2) DG eliminates the need for target data by learning representations across multiple source domains. In CSI-based systems, methods employ data augmentation (e.g., domain randomization in WiSR \cite{liu2023wisr}), attention mechanisms (e.g., Wigrunt \cite{gu2022wigrunt}), or physical propagation models (e.g., extracting Doppler shifts in Widar 3.0 \cite{qian2017widar} and WiHF \cite{li2020wihf}). Despite their success, traditional DG methods share a critical bottleneck: they rely heavily on massive, fully labeled source datasets, which are prohibitively expensive to acquire for non-interpretable CSI signals.

SSDG represents a more practical yet underexplored paradigm, designed for scenarios where only a small fraction of source data is labeled alongside abundant unlabeled data. 
% While SSDG has recently gained traction in the computer vision community—with methods like StyleMatch \cite{zhou2023semi} employing multi-view consistency and FBC-SA \cite{galappaththige2024towards} aligning pseudo-labels—it remains largely unaddressed in Wi-Fi sensing. 
To the best of our knowledge, the proposed \textbf{ARC-Fi} is the first dedicated SSDG framework for CSI data. 
%It is distinguished from image-based SSDG methods by its domain-specific physical insight, explicitly preventing paradigm-specific environmental shortcuts.

\subsection{Contrastive Representation Learning in Wi-Fi Sensing}
While contrastive learning \cite{chen2020simple,he2020momentum} excels in various domains, directly applying it to raw CSI often fails due to severe environmental confounders. To bridge this modality gap, recent studies inject domain-specific priors via heuristic augmentations, such as scalograms \cite{saeed2020federated} or Doppler traces \cite{lyons2024wifiact}. However, these engineered mathematical transformations frequently trigger the ``augmentation dilemma.'' Instead of learning robust gesture dynamics, models exploit these artificial transformations as superficial shortcuts, memorizing device-specific artifacts that cause catastrophic failures in downstream cross-domain tasks \cite{li2022unsupervised,song2024unleashing}.

In stark contrast to these spectrogram-based approaches, \textbf{ARC-Fi} operates directly on time-domain CSI and dismantles shortcut learning from a fundamental physical perspective. Our ARC module exploits the inherent spatial diversity across co-located antennas. Because this variance stems from distinct physical propagation paths of the same macroscopic event, it serves as a natural, semantics-preserving augmentation. This hard-to-forge physical variance strictly prohibits superficial memorization, compelling the model to distill genuine, domain-invariant gesture representations that are highly beneficial for label-efficient SSDG tasks.

\section{Method}

\label{sec:Method}
\begin{figure*}[!t]
	\centering
	\includegraphics[scale=0.54]{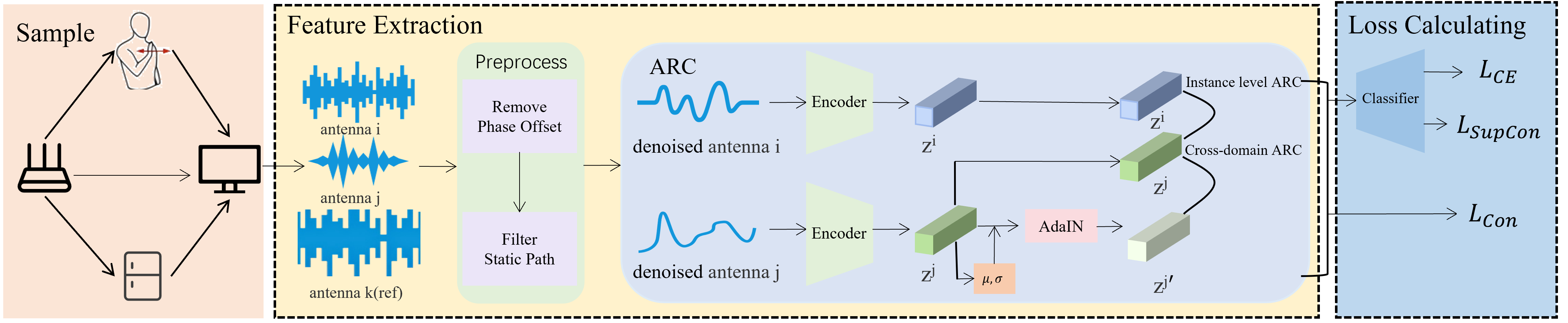}
	\caption{Overview of the proposed \textbf{ARC-Fi} framework. 
    The upper part of the figure presents the end-to-end workflow. Each CSI sample first undergoes preprocessing to remove phase offset and static path, mitigating the shift in $p(y|h)$. 
    The processed signals from different antennas are treated as natural intra-domain augmented views, while one antenna’s sample is adaptively normalized to generate a cross-domain ARC-augmented view.
    All anchor and augmented samples are then fed into the loss module for joint optimization. The overall objective integrates supervised and unsupervised contrastive losses ($L_{SupCon}$ and $L_{Con}$) in a weighted form, together with a cross-entropy ($L_{CE}$) term for source-domain classification, to reduce undesired repulsion between same-class samples across domains.
    } 
	\label{fig:overview}
\end{figure*}

\subsection{Overview and Preliminary}
As illustrated in Figure ~\ref{fig:overview}, \textbf{ARC-Fi} comprises four key components: a preprocessing stage, a feature extraction module, an ARC-based enhanced view construction module, and the integration of supervised loss for labeled data alongside self-supervised loss for unlabeled data.

\paragraph{CSI Data Format}
Before detailing these specific modules, we we would like to provide a brief introduction to the format of CSI data used in this paper.
Wi-Fi CSI is represented as a four-dimensional tensor, denoted as:
\begin{align}
H = \{h_1,h_2,...,h_B\} \in \mathbb{C}^{B \times M \times K \times T}.
\end{align}
In our notation $B$, $M$, $K$, and $T$ correspond to the number of data samples, receiver antennas, transmit antennas, subcarriers, and packets, respectively.
We omit the number of transmitter as the public datasets used in this paper has only one antenna.

\paragraph{SSDG Problem Setting}
The source domain comprises $N$ distinct sub-domains (e.g., different rooms), denoted as $\mathcal{S} = \bigcup_{n=1}^{N} \mathcal{S}_n$. Each sub-domain $\mathcal{S}_n$ is partitioned into a small labeled subset $\mathcal{S}_n^L = \{(h_{n,i}^L, y_{n,i}^L)\}_{i=1}^{B_n^L}$ and an abundant unlabeled subset $\mathcal{S}_n^U = \{h_{n,j}^U\}_{j=1}^{B_n^U}$. Here, $h$ denotes the raw CSI input, $y$ is the gesture label, and $B_n^L \ll B_n^U$. 
The target domain is defined as $\mathcal{T} = \{h_k^\mathcal{T}\}_{k=1}^{B^\mathcal{T}}$. Crucially, $\mathcal{T}$ remains strictly inaccessible during the entire training phase.

% Our objective is to leverage the partially labeled source data $\mathcal{S}$ to learn a domain-generalizable model $F = g \circ f$ that accurately predicts labels on the out-of-distribution target data $\mathcal{T}$. Specifically, the feature encoder $f: \mathcal{H} \rightarrow \mathcal{Z}$ maps the input CSI space to a domain-agnostic latent space, and the classifier $g: \mathcal{Z} \rightarrow \mathcal{Y}$ outputs the gesture category, establishing a robust end-to-end mapping.
% \paragraph{SSDG Problem Setting}
% The source domain comprises $N$ distinct sub-domains (e.g., different rooms), denoted as $\mathcal{S} = \bigcup_{n=1}^{N} \mathcal{S}_n$. Under the SSDG constraint, each $\mathcal{S}_n$ contains a limited labeled subset $\mathcal{S}_n^L = \{(h_i, y_i)\}_{i=1}^{B_n^L}$ and an abundant unlabeled subset $\mathcal{S}_n^U = \{h_j\}_{j=1}^{B_n^U}$, where $B_n^L \ll B_n^U$. Our objective is to train a model $F = g \circ f$ exclusively on the partially labeled source data $\mathcal{S}$ that accurately generalizes to an unseen, strictly inaccessible target domain $\mathcal{T}$. Specifically, the encoder $f: \mathcal{H} \rightarrow \mathcal{Z}$ distills domain-agnostic features from the CSI space, and the classifier $g: \mathcal{Z} \rightarrow \mathcal{Y}$ outputs the final gesture predictions.

\subsection{The Domain Shift of CSI Data}
\label{sec:domainshift}
According to former studies \cite{li2017indotrack}, CSI data can be described as following formula:
\begin{align}
H(f,t) &= e^{-j\theta} \big( H_s(f,t) + H_d(f,t) \big)  \notag \\
&= \underbrace{e^{-j\theta}}_{\text{phase offset}} 
\big( \underbrace{H_s(f,t)}_{\text{static path}} 
+ \underbrace{A(f,t) e^{-j2\pi \frac{d(t)}{\lambda}}}_{\text{dynamic path}}  
\big) .
\end{align}

This formulation indicates that CSI comprises three components: a phase offset, a static path component, and a dynamic path component. 
The phase offset is a time-varying random noise induced primarily by device hardware imperfections. 
It is generally treated as noise and is typically mitigated using preprocessing techniques such as the CSI ratio method \cite{wu2022wifi} or CSI conjugate multiplication \cite{li2017indotrack}.

From a probabilistic standpoint, cross-domain shifts in CSI sensing stem from two distinct and independent sources:
\begin{itemize}
    \item Environmental Confounding (Shift in $p(y|h)$): The environment's static multipath profile acts as a confounder, analogous to an image background. It is causally unrelated to the gesture ($y$) but creates spurious correlations in the raw signal ($h$). A change in environment alters $p(y|h)$, causing models that learn these spurious features to fail.

    \item Covariate Shift (Shift in $p(h)$): Variations in user location, body type, and habits induce a covariate shift in the dynamic path, altering the input gesture distribution $p(h)$. The core challenge is to learn a representation that is invariant to these shifts while preserving the causal link between motion and the gesture label.
\end{itemize}

% Existing methods typically fail because they conflate these two distinct problems, applying a single, monolithic learning strategy to the composite signal. This forces the model to learn a brittle solution that cannot disentangle causal features from environmental artifacts.

% Our core insight is that these two shifts must be decoupled and addressed sequentially. We first use targeted signal processing to eliminate the static path, thereby removing the environmental confounder and stabilizing $p(y|h)$. Only then do we apply deep learning to the cleaned signal to learn a representation that is robust to the remaining covariate shifts in $p(h)$.
Existing methods typically fail by conflating these two shifts, yielding brittle models that entangle causal features with environmental artifacts. Our core insight is to decouple them: we first eliminate the static path via targeted signal processing to stabilize $p(y|h)$, and subsequently apply deep learning to the purified signal to learn representations robust to covariate shifts in $p(h)$.

\subsection{Pre-processing}
Our pre-processing systematically purifies the raw CSI by mitigating hardware noise and environmental static paths.

\paragraph{Phase Offset Mitigation}
As mentioned above, we adopted a CSI ratio module to eliminate phase offset:
\begin{align}
H_r(f,t) &= \frac{H^{(i)}(f,t)}{H^{(j)}(f,t)} 
% &= \frac{e^{-j\theta} \big( H_s^{(1)} + A^{(1)} e^{-j2\pi \frac{d^{(1)}(t)}{\lambda}}\big)}{e^{-j\theta} \big( H_s^{(2)} + A^{(2)} e^{-j2\pi \frac{d^{(2)}(t)}{\lambda}} \big)}  \\
% &= \frac{ H_s^{(1)} + A^{(1)} e^{-j2\pi \frac{d^{(2)}(t)+\Delta d}{\lambda}}}{H_s^{(2)} + A^{(2)} e^{-j2\pi \frac{d^{(2)}(t)}{\lambda}}} \notag 
\end{align}
where $H^{(i)}(f,t)$ and $H^{(j)}(f,t)$ are the CSI of two receiver antennas. Instead of randomly selecting, we adopt the antenna selection coefficient proposed in \cite{zhang2025wiopen} to determine the optimal reference antenna $H^{(j)}(f,t)$. This coefficient is computed as follows:

\begin{align}
s_a = \frac{1}{K}\sum_{k=1}^K\frac{var(|H_a(f_k,t)|)}{mean(|H_a(f_k,t)|)}.
\end{align}
We choose the antenna with the lowest $s_a$ as reference antenna.

\paragraph{Static Path Suppression}
To remove the environmental confounder, we apply Time-Domain Differential CSI (TD-CSI) \cite{10.1145/3659608} for datasets with timestamp information. For datasets lacking explicit temporal data, we achieve an equivalent suppression using Low-Cut Filtering (LCF).
% To remove the environmental confounder, we apply Time-Domain Differential CSI (TD-CSI) \cite{10.1145/3659608} for datasets with timestamp information. For datasets lacking explicit temporal data, we achieve an equivalent suppression using Low-Cut Filtering (LCF).
% Subsequently, we incorporate a static path suppression step during preprocessing. Specifically, our strategy varies depending on the characteristics of each dataset. Prior research has demonstrated that employing time-domain differential CSI (TD-CSI) \cite{10.1145/3659608} achieves superior performance in removing static multipath components.
% In particular, TD-CSI computes the difference between two CSI measurements obtained within a short interval $\Delta t$. 
% However, the successful implementation of TD-CSI depends on the availability of timestamp information within the dataset. Accordingly, for datasets that include timestamp data, we utilize TD-CSI to suppress static paths. Conversely, for datasets lacking explicit temporal information, we adopt a low-cut filtering (LCF) to attenuate static components, thereby achieving a similar outcome.

\paragraph{Tensor Construction}
The filtered data $D_r$ is decomposed into amplitude $a$ (normalized to $[0, \pi]$) and phase $p$. 
These are concatenated to form the input tensor $x = [a, p] \in \mathbb{R}^{M \times 2K \times T}$. 
Finally, timestamp-based resampling is applied to ensure a uniform packet count $T$ across all samples for batch processing.

\begin{figure*}[!t]
	\centering
	\includegraphics[scale=0.07]{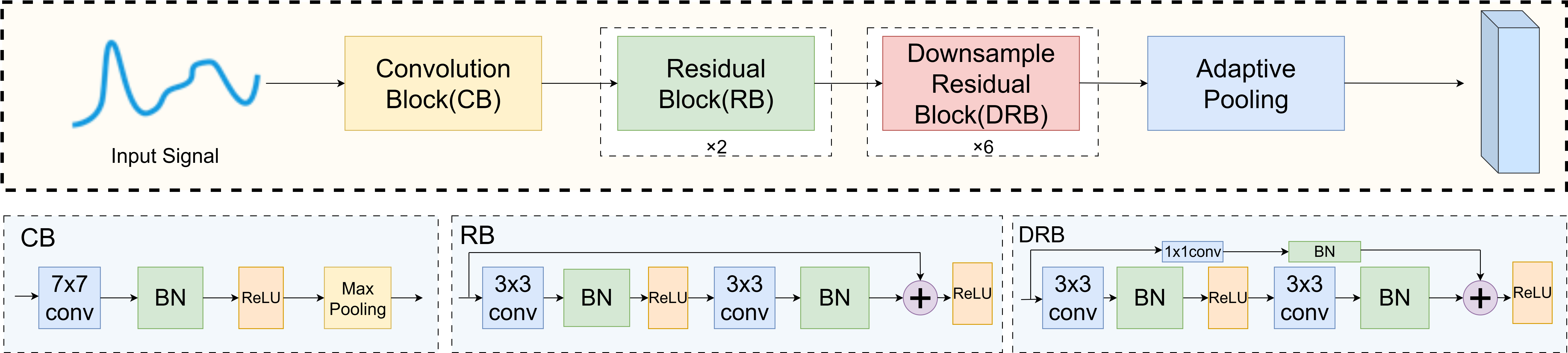}
	\caption{Overview of the proposed network architecture. The model takes the input signal and processes it sequentially through four main components: a convolutional block (CB), a residual block (RB), a downsample residual block (DRB), and an adaptive pooling layer. The CB consists of a 7×7 convolutional layer followed by batch normalization (BN), a ReLU activation, and a max pooling layer. The RB follows a standard residual structure with two consecutive 3×3 convolutional layers, each followed by BN and ReLU, and a residual shortcut connection. The DRB extends the RB by introducing an additional 1×1 convolution and BN in the residual path to enable downsampling.}
	\label{fig:TMC_Resnet}
\end{figure*}

\subsection{Network and Feature extraction}
To fully exploit antenna spatial diversity, we partition the preprocessed input $x$ along the antenna dimension $M$. Each antenna's data segment $x^{(i)} \in \mathbb{R}^{2K \times T}$ is independently processed by a shared encoder $F$ to extract latent representations:

% Following the preprocessing stage, the input data $x$ is partitioned along its antenna dimension $M$. Each resulting segment $x^{(i)}$ is then independently processed by an encoder $F$ to generate latent features $z^{(i)}$:
\begin{align}
z^{(i)} = F(x^{(i)}), \quad x^{(i)} \in \mathbb{R}^{2K \times T}, \quad i = 1, 2, \dots, M.
\end{align}
As illustrated in Figure ~\ref{fig:TMC_Resnet}, $F$ is a 2D Residual Network (ResNet) \cite{he2016deep, liu2023wisr} tailored for our multidimensional input.

% The encoder begins with an initial sequence of layers: a 2D convolutional layer (Conv2D), followed by batch normalization (BatchNorm2D), a Rectified Linear Unit (ReLU) activation function, and a max pooling layer (MaxPool2D). 
% The output is then fed through four sequential ResNet blocks. The architecture concludes with a final sequence comprising a Conv2D layer, BatchNorm2D, ReLU, and an adaptive average pooling layer to produce the feature representation $z^{(i)}$.

Consistent with standard contrastive learning frameworks, we employ a projection head $\mathrm{g}(\cdot)$ to map the extracted features into an embedding space suitable for contrastive loss computation. Specifically, the features $z^{(i)}$ are transformed by a two-layer multi-layer perceptron (MLP) to obtain the final embeddings $u^{(i)}$:
\begin{align}
u^{(i)} = G(z^{(i)}).
\end{align}

For the classification task, each classifier head $\phi$ consists of a single fully-connected layer that takes the feature representations $z^{(i)}$ as input to produce the final predictions.

\subsection{Antenna Response Consistency Module}
\label{ARC}
As mentioned in Section~\ref{sec:intro}, existing augmentation strategies may unintentionally introduce false positive views.
To overcome these limitations, we design the ARC module, which generates semantics-preserving positive views by exploiting the inherent physics of multi-antenna systems. 
ARC generates two complementary types of positive views: Natural Multi-Views and Cross-Domain Synthetic Views.

\subsubsection{Natural ARC}
In a typical multiple-input multiple-output (MIMO) system, although receiver antennas are co-located, minor hardware variations, spatial displacements, and polarization differences cause them to capture CSI signals with unique, task-irrelevant features, which we term "styles".
%, as illustrated in Figure~\ref{fig:ant_diff}. 
Crucially, these signals represent diverse perspectives of the same physical event and thus share a common, task-relevant core pattern, or "content." 
Building on this insight, we treat embedding from different antennas $u^{(i)}$ and $u^{(j)}$ of the same sample as a set of natural, implicitly augmented positive pairs.
% If we use Equation \ref{eq:dr} to describe $D_r^{(i)}$, then $D_r^{(j)}$ can be described in following format:
% \begin{align}
% D_r^{(j)}
% &= 2A \sin{\frac{\Delta (\theta (t) +\theta_{\Delta d})}{2}e^{j((\theta (t) +\theta_{\Delta d})+\frac{\Delta (\theta (t) +\theta_{\Delta d})}{2}+\frac{\pi}{2})}}
% \end{align}
% The phase difference $\theta_{\Delta d}$ arises from the spatial separation $\Delta d$ between antennas $i$ and $j$, and can be expressed as $-2\pi \frac{\Delta d}{\lambda}$
% This phase difference serves as a natural data augmentation stemming from the spatial diversity of antennas.
Contrasting between these views forces the model to learn features that are invariant to the intrinsic stylistic differences between antennas, providing a baseline for robust representation learning.

% 在这里加一张图用来说明adain地作用 TODO 
\subsubsection{Cross-Domain Antenna Style Transfer}
While natural views guarantee semantic consistency, their diversity is limited by the fixed characteristics of the hardware. To train a model capable of generalizing to a much wider spectrum of unseen domain shifts, we introduce a generative strategy to synthesize more challenging positive samples. For this, we add an adaptive instance normalization (AdaIN) \cite{huang2017adain} operator.
Given the feature representations of two antenna signals from different domains, $u^{\text{AdaIN}} = \text{AdaIN}(u^{(i)},u^{(j)})$, the AdaIN operation is defined as:
\begin{align}
    u^{\text{AdaIN}} = \sigma(u^{(i)}) \left (\frac{u^{(j)}-\mu \left (u^{(j)} \right)}{\sigma{\left (u^{(j)} \right)}} \right)+\mu{(u^{(i)})},
\end{align}

where $\mu(\cdot)$ and $\sigma(\cdot)$ are the mean and standard deviation of the feature map. 
This operation retains the core activity "content" from antenna $i$ while explicitly injecting the response "style" of antenna $j$. 
This elevates a simple multi-view comparison into a sophisticated mechanism for compelling the model to learn features that are truly invariant to cross-domain variations.

\emph{The Difference Between ARC and CSI Ratio:}
We would like to emphasize that our proposed ARC is\textbf{ fundamentally different from the widely used CSI ratio}, although the two can easily be confused. In short, the CSI ratio leverages clock synchronization between different antennas on the same device to eliminate phase offset, whereas ARC exploits the spatial diversity between antennas to construct a data augmentation strategy — building on top of the CSI ratio preprocessing. 

\subsection{Unified Semi-Supervised Contrastive Objective}
To fully exploit the cross-domain views generated by the ARC module and prevent class misalignment, we propose a unified contrastive objective that seamlessly integrates unsupervised and pseudo-label-guided supervised losses.

\textbf{Unsupervised Contrastive Loss ($\mathcal{L}_{\text{UCon}}$).}
Based on the NT-Xent loss \cite{chen2020simple}, $\mathcal{L}_{\text{UCon}}$ pulls together the anchor $u_q^{(i)}$ with its inter-antenna positive view $u_q^{(j)}$ and the AdaIN-synthesized cross-domain view $u_q^{\text{AdaIN}}$:
\begin{align}
\mathcal{L}_{\text{UCon}} &= -\sum_{q=1}^B \Bigg( \log \frac{\exp(\text{sim}(u_q^{(i)},u_q^{(j)})/\tau)}{\sum_{r \in O \setminus \{u_q^{(i)}\}} \exp(\text{sim}(u_q^{(i)},u_r)/\tau)} \notag \\
&\quad + \log \frac{\exp(\text{sim}(u_q^{(i)},u_q^{\text{AdaIN}})/\tau)}{\sum_{r \in O \setminus \{u_q^{(i)}\}} \exp(\text{sim}(u_q^{(i)},u_r)/\tau)} \Bigg),
\end{align}
where $B$ is the batch size, $O$ denotes all augmented views, $\text{sim}(\cdot, \cdot)$ is the cosine similarity, and $\tau$ is the temperature parameter. 

\textbf{Supervised Contrastive Loss ($\mathcal{L}_{\text{SCon}}$).}
Relying solely on $\mathcal{L}_{\text{UCon}}$ risks repelling cross-domain negative pairs that actually share the same gesture semantics. To explicitly align domain-invariant features, we extend the supervised contrastive loss \cite{khosla2020supervised} to leverage both scarce ground-truth labels and abundant pseudo-labels:
\begin{align}
\mathcal{L}_{\text{SCon}} &= \sum_{q=1}^B \frac{-1}{|P(q)|} \sum_{p\in P(q)} \notag \\
&\quad \times \log \frac{\exp(\text{sim}(u_q^{(i)},u_p)/\tau)}{\sum_{r \in O \setminus \{u_q^{(i)}\}} \exp(\text{sim}(u_q^{(i)},u_r)/\tau)},
\end{align}
where $P(q) = \{p \in O \setminus \{u_q^{(i)}\} : y_p = y_q \text{ or } \hat{y}_p = y_q\}$ is the positive set containing samples that share the anchor's label. Here, $\hat{y}_p$ denotes the pseudo-label for unlabeled data, which is refined via the cross-domain consistency regularizer proposed in \cite{Gal2024FBCSA} to ensure reliability. $|P(q)|$ represents the cardinality of this set.

\textbf{Overall Objective.}
To ensure the learned feature representations are strictly domain-agnostic while remaining discriminative for the downstream classification task, we combine the unified contrastive loss with a standard cross-entropy (CE) loss. The final training objective is formulated as:
\begin{align}
\mathcal{L} = \alpha \mathcal{L}_{\text{SCon}} + (1-\alpha)\mathcal{L}_{\text{UCon}} + \mathcal{L}_{\text{CE}},
\end{align}
where $\alpha \in (0, 1)$ is a hyperparameter balancing the supervised and unsupervised contrastive components.

\section{Experiments}
% In this section, we present the experimental results for both DG and SSDG. We further compare our proposed method with state-of-the-art Unsupervised Domain Adaptation (UDA) approaches. Finally, we conduct an ablation study to analyze the contribution of each component in our framework.
\label{sec:Experiments}

\subsection{Datasets and Evaluation Protocol}
We evaluate our method on two public gesture datasets: Widar \cite{qian2017widar} and CSIDA \cite{zhang2021wifi}. Selected for their distinct hardware configurations (Intel vs. Atheros NICs), subcarrier dimensions, and gesture vocabularies, these datasets provide a rigorous testbed. To comprehensively evaluate \textbf{ARC-Fi} while ensuring statistical validity, we assign specific evaluation roles to each dataset based on their scale and hardware characteristics.
\textbf{a) Widar} \cite{qian2017widar} comprises recordings from 17 users engaging in 22 gesture activities across 3 distinct environments. Data was collected using Intel 5300 NICs (1 transmitter with 1 antenna, 6 receivers with 3 antennas each) at a sampling rate of 1000 packets per second, yielding compressed CSI data with 30 subcarriers per link. For our evaluation, we utilize data from a single receiver across all rooms. To standardize the naturally variable sequence lengths, all samples are temporally downsampled to a fixed length of $T=800$. Given its massive scale and spatial diversity, Widar serves as our primary benchmark for evaluating the highly challenging Semi-Supervised Domain Generalization (SSDG) setting.
\textbf{b) CSIDA} \cite{zhang2021wifi} contains 6 gestures performed by 5 users (3 males, 2 females) across 2 distinct rooms. Data was collected using Atheros NICs \cite{Xie:2015:PPD:2789168.2790124} with a 1-transmitter and 1-receiver setup (the receiver is equipped with 3 antennas). This configuration captures uncompressed CSI at a rate of 1000 packets per second, yielding 114 subcarriers per link. Each gesture sample has a strictly fixed temporal duration of 1.8 seconds. Because its relatively limited data volume cannot support a statistically robust SSDG partition without severe sampling bias, we specifically employ CSIDA under the standard Domain Generalization (DG) setting. This deliberate protocol allows us to independently validate that our core physics-informed ARC module remains highly effective across different underlying Wi-Fi protocols and hardware architectures, even without the aid of semi-supervised optimization.

\subsection{Baseline}
% \begin{table}[htbp]
% \centering
% \renewcommand{\arraystretch}{1}  
% \caption{Key Differentiators of \textbf{ARC-Fi} from Baseline Methods.}
% \label{tab:baseline}
% \setlength{\tabcolsep}{3pt} % 默认6pt，这里改小
% \begin{tabular}{cccc}
% \toprule
% \textbf{Method} & \textbf{CSI-Specific} & \textbf{Cross-Domain} & \textbf{Use Unlabeled Data}  \\
% \midrule
% WiSDA \cite{jiao2024wisda}  & \cmark  & UDA & \xmark  \\ 
% \addlinespace % Adds a bit of visual separation between rows
% CoTMix \cite{eldele2023contrastive} & \xmark   & UDA & \xmark \\
% \addlinespace
% AdvSKM \cite{liu2021adversarial} & \xmark   & UDA & \xmark \\
% \addlinespace
% WiSR \cite{liu2023wisr}   & \cmark  & DG  & \xmark  \\
% \addlinespace
% WiOpen \cite{zhang2025wiopen} & \cmark  & DG  & \xmark  \\
% \addlinespace
% WiGrunt  \cite{gu2022wigrunt} & \cmark   & DG  & \xmark  \\
% \addlinespace
% ERM  \cite{vapnik1991principles} & \xmark   & DG  & \xmark  \\
% \addlinespace
% SimCLR \cite{chen2020simple} & \xmark   & Not Cross-Domain & \cmark  \\
% \midrule
% % \textbf{ARC-Fi}  & Yes   & UDA \& DG \& SSDG & Yes  \\
% % \bottomrule
% \textbf{ARC-Fi}  & \cmark   & SSDG & \cmark  \\
% \bottomrule
% \end{tabular}
% \end{table}

As the first dedicated SSDG framework for Wi-Fi sensing, \textbf{ARC-Fi} lacks direct CSI-specific SSDG baselines. To ensure a rigorous evaluation, we compare it against eight representative algorithms from related cross-domain paradigms, categorized as follows:
\textbf{1) UDA Methods} (require unlabeled target data): Includes \textit{WiSDA} \cite{jiao2024wisda} (CSI-to-image augmentation with cosine alignment), \textit{CoTMix} \cite{eldele2023contrastive} (temporal mixup and contrastive learning), and \textit{AdvSKM} \cite{liu2021adversarial} (adversarial spectral kernel matching).
\textbf{2) DG Methods} (rely on fully labeled source data): Comprises \textit{WiSR} \cite{liu2023wisr} (domain randomization and adversarial learning), \textit{WiOpen} \cite{zhang2025wiopen} (uncertainty-aware dynamic boundaries), and \textit{WiGrunt} \cite{gu2022wigrunt} (spatial-temporal dual-attention).
\textbf{3) Foundational Baselines} (for performance collapse analysis): We evaluate \textit{ERM} \cite{vapnik1991principles} (the standard supervised lower bound) and \textit{SimCLR} \cite{chen2020simple} (a generic vision-inspired contrastive framework). These are explicitly included to demonstrate a critical vulnerability: naively porting standard machine learning paradigms without accounting for underlying RF physical properties inevitably triggers severe shortcut learning and catastrophic performance collapse in unseen environments.

% \subsection{Implementation Detail}
% % 训练的学习率等超参数
% All experiments were conducted using Nvidia A6000 GPUs. Model performance was evaluated using accuracy.
% In the cross-domain experiments, we consider multiple source domains and one target domain, each corresponding to different room, location, orientation or user settings.
% For the source domains, we split the data into a training set and a validation set using an 8:2 ratio.

\begin{figure*}[!tbh]
    \label{fig:semidg}
    \centering
    \subfloat[Room]{\includegraphics[width=0.5\textwidth]{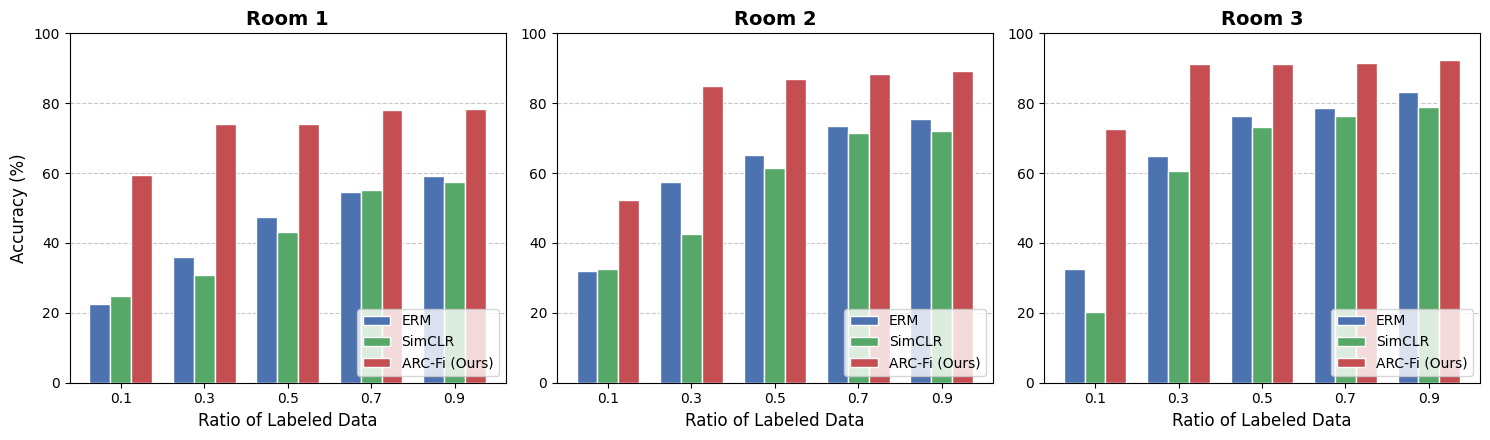}\label{fig:Room}}
    \subfloat[Location]{\includegraphics[width=0.5\textwidth]{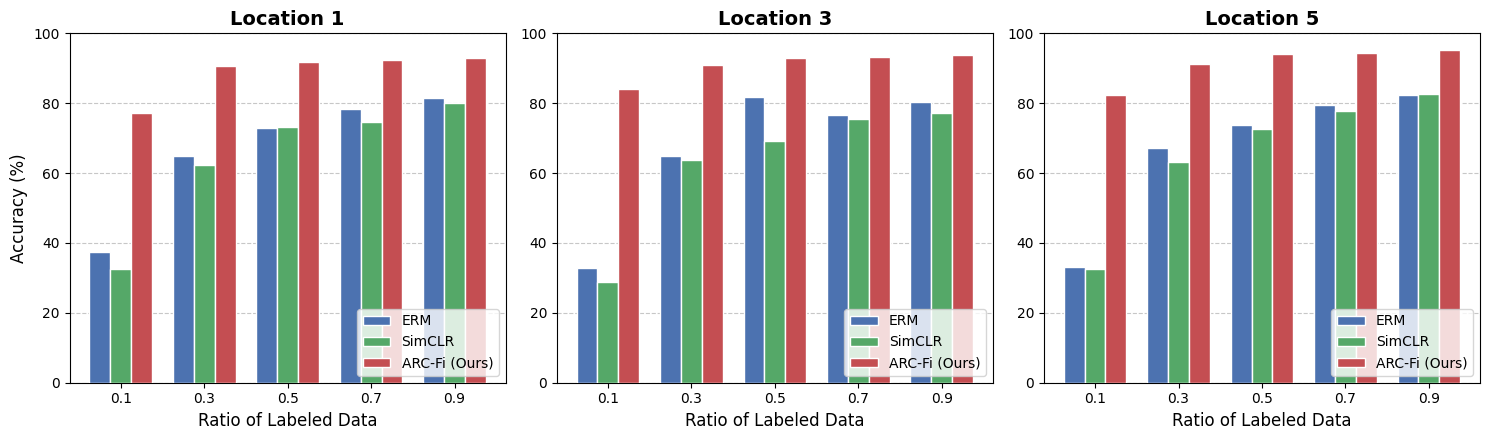}\label{fig:Loc}}\\
    \subfloat[User]{\includegraphics[width=0.5\textwidth]{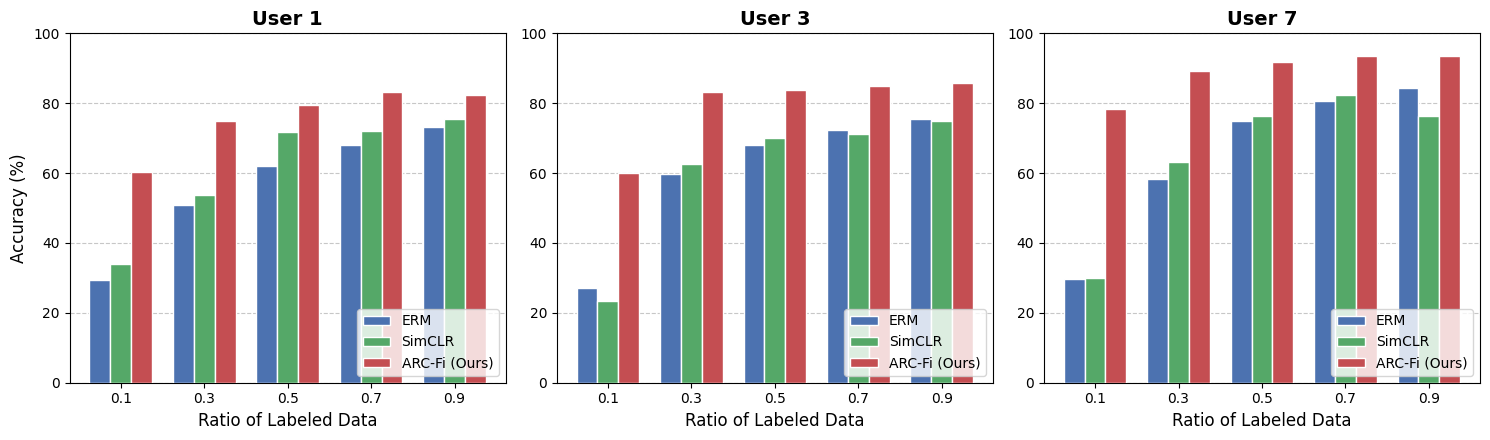}\label{fig:User}}
    \subfloat[Orientation]{\includegraphics[width=0.5\textwidth]{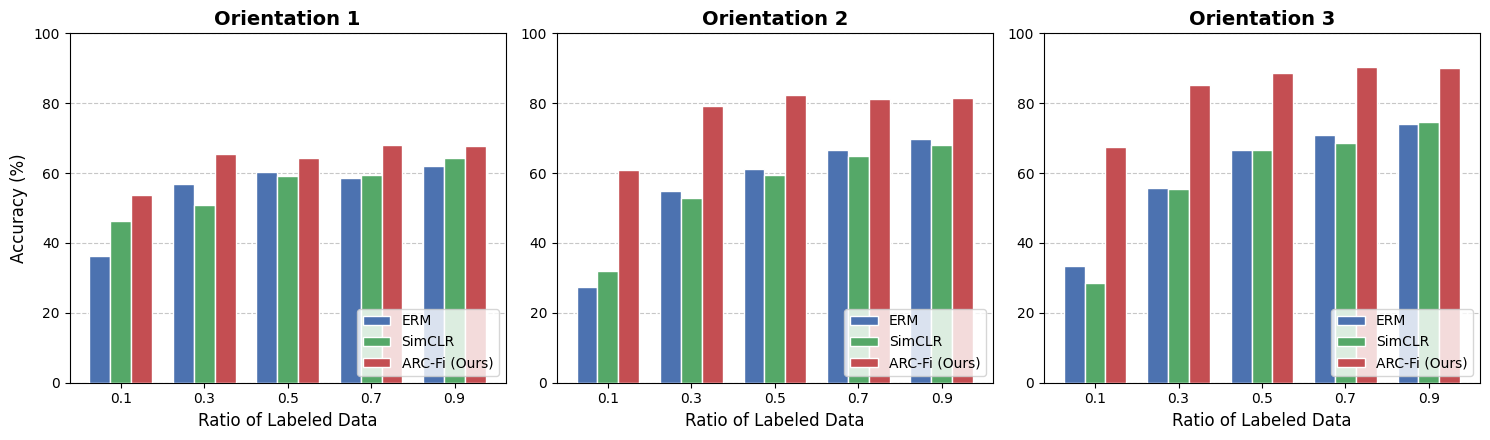}\label{fig:Ori}}
    \caption{The SSDG performance on Widar dataset for ERM, SimCLR and \textbf{ARC-Fi}. }
\end{figure*}

\subsection{Analysis of SSDG Performance}

We evaluate data efficiency under a SSDG setting, where models are trained on a source domain with varying ratios of labeled data (from a sparse 10\% to 90\%) and tested on an unseen target domain.

The results, shown in Fig.~\ref{fig:semidg}, are unequivocal. \textbf{ARC-Fi} decisively outperforms all baselines across every label ratio. This advantage is most pronounced in the low-data regime—the most critical scenario for practical deployments. With only 10\% of labels, \textbf{ARC-Fi} delivers an absolute performance gain of 20-50 percentage points over SimCLR. This margin, while narrowing with more supervision, remains substantial even at 90\% labels (6\%). This trend confirms our framework's exceptional ability to extract robust representations from unlabeled data when supervision is scarce.

The baseline failures are illuminating. As expected, ERM collapses from overfitting when labels are sparse. More telling is SimCLR's performance. Its reliance on generic, vision-based augmentations encourages shortcut learning, where the model learns superficial augmentation artifacts instead of the gesture's core physical semantics. This makes the resulting representations brittle and ineffective without strong supervision.

\textbf{ARC-Fi}, in contrast, is fundamentally engineered for this challenge. By grounding the learning process in physics-informed priors, our framework extracts meaningful and generalizable features from the unlabeled data stream. This enables it to build a robust model with minimal labeled examples, establishing its viability for practical, large-scale deployments where annotated data is a luxury.

\begin{figure}[!htbp]
    \centering
    \subfloat[WiSDA]{%
        \includegraphics[width=0.242\textwidth]{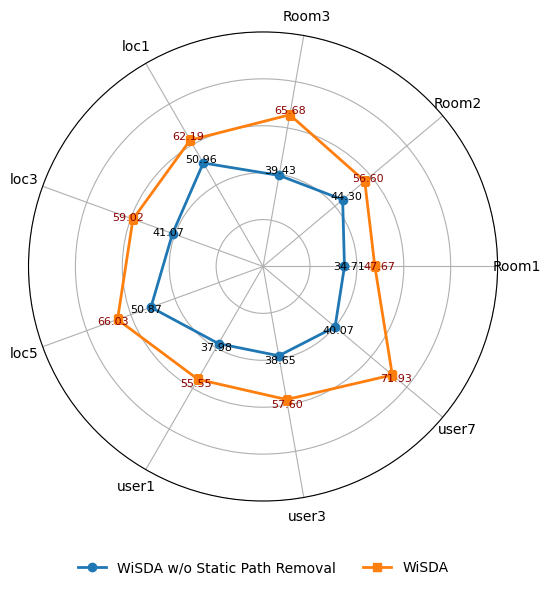}%
        \label{fig:wisda}%
    }
    \hfill
    \subfloat[COTMIX]{%
        \includegraphics[width=0.242\textwidth]{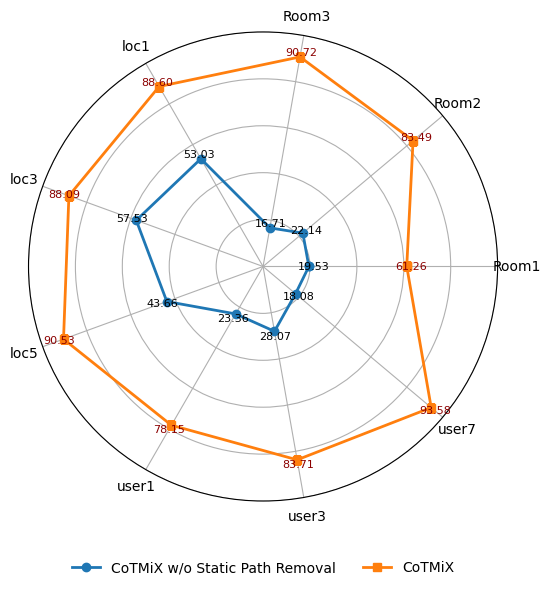}%
        \label{fig:cotmix}%
    }
    \hfill
    \subfloat[AdvSKM]{%
        \includegraphics[width=0.242\textwidth]{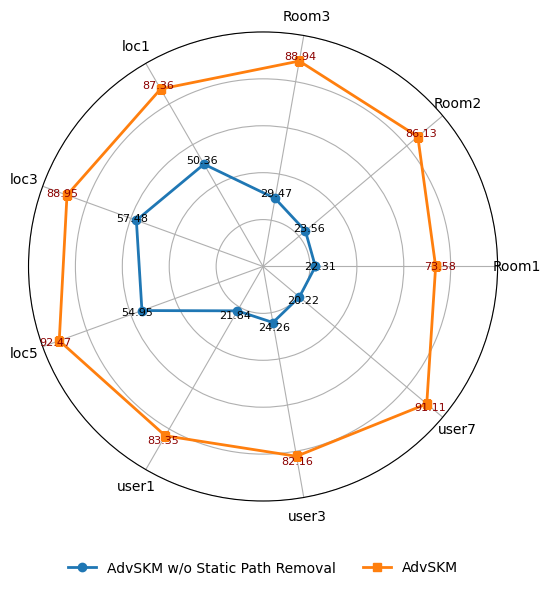}%
        \label{fig:advskm}%
    }
    \hfill
    \subfloat[Ours]{%
        \includegraphics[width=0.242\textwidth]{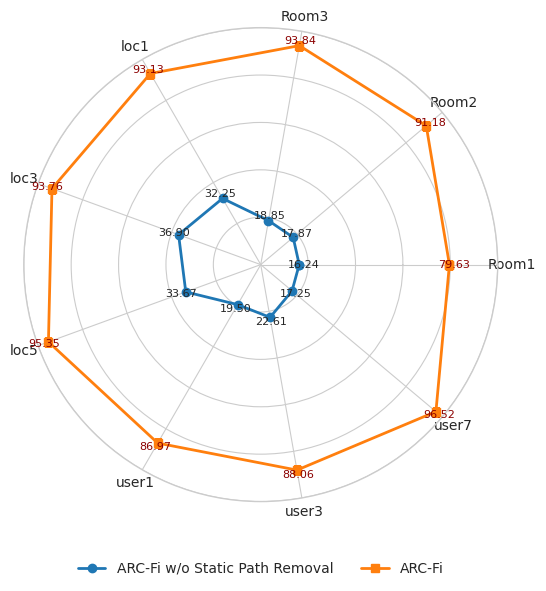}%
        \label{fig:ours}%
    }
    \caption{Performance comparison of WiSDA, COTMIX, AdvSKM, and our method with and without Static Path Removal on Widar cross-domain tasks under the domain generalization setting.}
    \label{fig:ablation_tdcsi}
\end{figure}

\subsection{Analysis of DG Performance}

\begin{table*}[!tbh] 
    \caption{
    DG accuracy (\%) of different algorithms on the Widar dataset.
    The rows are organized by cross-domain type (Room, Location, User and Orientation), with each target domain listed under its corresponding group.
    Each cell reports the mean accuracy in the target domain together with the standard deviation.
    The best performance in each row is highlighted in bold.
    To ensure a fair comparison, pre-processing pipelines are tailored to the algorithms as in their original works. Specifically, WiGrunt and WiOpen employ \textit{phase offset removal (POR)} followed by a \textit{data-to-image visualization (Vis)} transformation \cite{zhang2025wiopen,gu2022wigrunt}. All other methods, including ours, use a pipeline of POR and then get the TD-CSI as input.
    }
    \setlength{\tabcolsep}{6pt}
    \centering
    \renewcommand{\arraystretch}{1.4}
    \begin{tabular}{c|c|c|ccclc}
    \toprule
    \multirow{2}{*}{Cross-domain Type} &  \multirow{2}{*}{\shortstack{Training Data \\ (Labeled Source (LS) )}}&Algorithms& ERM&WiSR&WiGrunt
&  WiOpen
&Ours \\
    \cmidrule(lr){3-8}
    &  &Pre-Processing& POR+TD-CSI& POR+TD-CSI& POR+Vis&  POR+Vis&POR+TD-CSI\\
    \midrule
    \multirow{3}{*}{Room}&  LS: Room 2,3&Room 1& 61.74$\pm$1.91& 65.40$\pm$1.60& 67.60$\pm$1.54&   62.38$\pm$0.71&\textbf{79.63$\pm$0.30}\\
 & LS: Room 1,3& Room 2& 76.99$\pm$1.21& 78.32$\pm$1.61& 80.47$\pm$1.11& 74.78$\pm$1.00&\textbf{91.18$\pm$0.50}\\
        &  LS: Room 1,2&Room 3& 84.13$\pm$0.84& 85.21$\pm$0.95& 84.86$\pm$0.96&   83.59$\pm$0.77&\textbf{93.84$\pm$0.46}\\
    \midrule
    \multirow{3}{*}{Location} 
        &  LS: Loc 3,5&Loc 1& 83.24$\pm$1.16& 83.82$\pm$0.97& 85.72$\pm$0.32&   85.49$\pm$0.57
&\textbf{93.13$\pm$0.71}
\\
        &  LS: Loc 1,5&Loc 3& 83.89$\pm$0.83& 83.56$\pm$0.15& 85.94$\pm$0.39&   85.71$\pm$0.67
&\textbf{93.76$\pm$0.37}
\\
        &  LS: Loc 1,3&Loc 5& 85.13$\pm$1.28& 86.96$\pm$0.55& 88.86$\pm$1.11&   89.44$\pm$0.68
&\textbf{95.35$\pm$0.19}
\\
    \midrule
    \multirow{3}{*}{User}&  LS: User 3,7&User 1& 74.46$\pm$2.00& 75.46$\pm$0.99& 73.58$\pm$1.04&   69.69$\pm$1.10
&\textbf{86.97$\pm$0.81}
\\
        &  LS: User 1,7&User 3& 78.78$\pm$0.30& 79.48$\pm$0.40& 78.28$\pm$0.90&   77.65$\pm$0.77
&\textbf{88.06$\pm$0.75}
\\
        &  LS: User 1,3&User 7& 85.38$\pm$1.00& 85.66$\pm$1.10& 87.03$\pm$0.30&   84.96$\pm$1.10
&\textbf{96.52$\pm$0.50}
\\
    \midrule
    \multirow{3}{*}{Orientation}&  LS: Ori 2,3 ULT: Ori 1&Ori 1
& 61.31$\pm$0.33& 55.64$\pm$3.03& \textbf{77.56$\pm$0.60}&   74.11$\pm$2.11&68.66$\pm$2.75\\
        &  LS: Ori 1,3 ULT: Ori 2&Ori 2
& 69.77$\pm$1.22& 65.34$\pm$1.85& 77.16$\pm$1.44&   76.71$\pm$0.76&\textbf{84.03$\pm$0.88}\\
        &  LS: Ori 1,2 ULT: Ori 3&Ori 3& 75.47$\pm$1.30& 69.04$\pm$2.00& 77.81$\pm$0.50&   78.03$\pm$0.45&\textbf{90.68$\pm$0.33}\\
    \bottomrule
    \end{tabular}
    \label{tab:dgwidar}
\end{table*}

The experimental results, detailed in Table~\ref{tab:dgwidar}, establish the definitive superiority of our \textbf{ARC-Fi} framework under the strict Domain Generalization (DG) setting on the Widar dataset.

Overall, our method consistently sets a new state-of-the-art, outperforming the strongest baseline across almost all cross-domain tasks. This leap in generalizability is most evident in the highly challenging cross-room scenarios. For instance, when generalizing to Room 1 (from Rooms 2 and 3) and Room 2 (from Rooms 1 and 3), \textbf{ARC-Fi} achieves 79.63\% and 91.18\% accuracy, respectively. This delivers a substantial absolute improvement of over 10\% to 12\% compared to the next-best approach (WiGrunt, achieving only 67.60\% and 80.47\% on the same tasks). This confirms that our physics-informed representations effectively capture domain-invariant gesture semantics rather than environment-specific background patterns.

Crucially, Table~\ref{tab:dgwidar} also reveals a fascinating anomaly that starkly exposes the dangers of shortcut learning in vision-inspired baselines. This fragility is best exemplified by the ``Orientation 1'' task. In this specific configuration, the user performs the gesture parallel to the transceiver's line-of-sight. According to the Fresnel Zone model, gestures moving in this parallel direction induce minimal signal reflection and dynamic phase modulation, making it physically the most difficult scenario for feature extraction. As expected, models that respect signal integrity, including our \textbf{ARC-Fi} and the foundational ERM baseline, exhibit a natural performance drop (yielding accuracies of 68.66\% and 61.31\%, respectively), correctly reflecting this severe physical information loss.

Paradoxically, the vision-based methods (WiGrunt and WiOpen) achieve unusually high accuracies of 77.56\% and 74.11\% on this physically degraded task. Rather than indicating superior gesture recognition, this ``high performance'' is a classic symptom of shortcut learning. Because their rigid data-to-image (POR+Vis) transformations destroy intrinsic RF phase semantics, these models are forced to overfit to spurious environmental artifacts or static background noise inadvertently captured during the Orientation 1 collection. This hypothesis is heavily corroborated by their catastrophic failure to generalize consistently across other physical domains (as later demonstrated in our cross-hardware analysis in Section~\ref{sec:RobustnessAcrossHdw}). By strictly anchoring its augmentations in RF physical laws, \textbf{ARC-Fi} completely avoids these superficial shortcuts, ensuring reliable and physically sound performance.

\subsection{Analysis of UDA Performance}

% \begin{table}[tbh]
%     \caption{
%     UDA accuracy (\%) on the Widar dataset across four domain shift settings.
%     Results are aggregated (Mean$\pm$Std) across all target sub-domains.
%     The specific pre-processing pipeline for each method is indicated directly below its name.
%     The best performance is highlighted in bold.
%     }
%     \label{tab:udawidar}
%     \centering
%     \renewcommand{\arraystretch}{1.2}
%     % 使用 resizebox 将表格宽度严格限制为单栏宽度 (\columnwidth)
%     \resizebox{\columnwidth}{!}{
%     \begin{tabular}{lcccc}
%     \toprule
%     \multirow{2}{*}{\textbf{Domain Shift}} & \textbf{WiSDA} & \textbf{CoTMIX} & \textbf{AdvSKM} & \textbf{ARC-Fi} \\
%     \cmidrule(lr){2-5}
%     & \textit{\footnotesize POR+TD-CSI} & \textit{\footnotesize POR+TD-CSI} & \textit{\footnotesize POR+TD-CSI} & \textit{\footnotesize POR+TD-CSI} \\
%     \midrule
%     Room        & 56.65$\pm$9.01 & 78.49$\pm$15.35 & 82.88$\pm$8.18  & \textbf{88.32$\pm$6.68} \\
%     Location    & 62.41$\pm$3.51 & 89.07$\pm$1.29  & 89.59$\pm$2.62  & \textbf{95.73$\pm$1.09} \\
%     User        & 61.69$\pm$8.92 & 85.15$\pm$7.81  & 85.54$\pm$4.86  & \textbf{90.85$\pm$4.04} \\
%     Orientation & 52.39$\pm$5.07 & 73.87$\pm$8.11  & 70.41$\pm$12.47 & \textbf{85.57$\pm$11.89} \\
%     \midrule
%     \textbf{Average} & 58.28 & 81.65 & 82.11 & \textbf{90.12} \\
%     \bottomrule
%     \end{tabular}
%     }
% \end{table}
\begin{table}[tbh]
    \caption{
    UDA accuracy (\%) on the Widar dataset across four domain shift settings.
    Results are aggregated (Mean$\pm$Std) across all target sub-domains.
    The specific pre-processing pipeline for each method is indicated directly below its name.
    The best performance is highlighted in bold.
    }
    \label{tab:udawidar}
    \centering
    \renewcommand{\arraystretch}{1.2}
    % 使用 resizebox 将表格宽度严格限制为单栏宽度 (\columnwidth)
    \resizebox{\columnwidth}{!}{
    \begin{tabular}{lcccc}
    \toprule
    \multirow{2}{*}{\textbf{Domain Shift}} & \textbf{WiSDA} & \textbf{CoTMIX} & \textbf{AdvSKM} & \textbf{ARC-Fi} \\
    \cmidrule(lr){2-5}
    & \textit{\footnotesize POR+TD-CSI} & \textit{\footnotesize POR+TD-CSI} & \textit{\footnotesize POR+TD-CSI} & \textit{\footnotesize POR+TD-CSI} \\
    \midrule
    Room        & 56.65$\pm$9.01 & 78.49$\pm$15.35 & 82.88$\pm$8.18  & \textbf{88.32$\pm$6.68} \\
    Location    & 62.41$\pm$3.51 & 89.07$\pm$1.29  & 89.59$\pm$2.62  & \textbf{95.73$\pm$1.09} \\
    User        & 61.69$\pm$8.92 & 85.15$\pm$7.81  & 85.54$\pm$4.86  & \textbf{90.85$\pm$4.04} \\
    Orientation & 52.39$\pm$5.07 & 73.87$\pm$8.11  & 70.41$\pm$12.47 & \textbf{85.57$\pm$11.89} \\
    \midrule
    \textbf{Average} & 58.28 & 81.65 & 82.11 & \textbf{90.12} \\
    \bottomrule
    \end{tabular}
    }
\end{table}

Even under the UDA setting, our proposed method exhibits remarkable generalization and robustness. As shown in Table~\ref{tab:udawidar}, \textbf{ARC-Fi} consistently achieves the highest classification accuracy across all four domain shift types. On the comprehensive Widar dataset, our model attains an average accuracy of 90.12\%, outperforming the strongest baseline, AdvSKM, by a substantial margin of over 8\%. This consistent, top-tier performance underscores the effectiveness of our model in adapting to diverse and unseen deployment conditions.

A critical advantage of \textbf{ARC-Fi} lies in its cross-domain stability. In our aggregated evaluation, the standard deviation reflects the performance fluctuation across different target sub-domains (e.g., varying target rooms or users). While competitive baselines like CoTMIX and AdvSKM achieve decent mean accuracies (often in the 80–89\% range), they suffer from severe performance degradation on specific target domains, resulting in high variance. For instance, under the highly challenging Room domain shift, CoTMIX exhibits a massive standard deviation of 15.35\%, indicating extreme instability across different rooms. In stark contrast, \textbf{ARC-Fi} restricts this fluctuation to just 6.68\% while simultaneously boosting the mean accuracy by nearly 10\%. Similarly, in the Location setting, our method yields a remarkably tight standard deviation of 1.09\%. Such stability indicates that our model’s success is not contingent on favorable target environments, but instead stems from learning genuinely robust, domain-invariant representations. This reliability is paramount for real-world off-the-shelf applications.

We attribute this markedly improved robustness to the proposed ARC module. Although CoTMIX also incorporates a contrastive objective, its augmentation strategy directly mixes source and target samples in the time domain. As argued in our augmentation dilemma, this heuristic approach risks destroying the physical semantic integrity of the RF signals, producing false-positive views that confuse the model in complex environments. By avoiding these pitfalls and grounding our augmentations in multi-antenna spatial physics, \textbf{ARC-Fi} consistently ensures superior and stable performance across all evaluated settings.

\subsection{Robustness Across Wi-Fi Hardware Architectures}
\label{sec:RobustnessAcrossHdw}
A critical concern in practical RF sensing is whether a model overfits to the specific artifacts of its underlying hardware. While our primary evaluations on the Widar dataset demonstrate \textbf{ARC-Fi}'s superiority on Intel 5300 NICs (which capture compressed CSI across 30 subcarriers), we leverage the CSIDA dataset to explicitly benchmark cross-hardware generalizability. Utilizing Atheros NICs to capture uncompressed CSI across 114 subcarriers, CSIDA presents a fundamentally different data distribution and sensing paradigm.

As detailed in Table~\ref{tab:dgcsida}, under the strict Domain Generalization (DG) setting, \textbf{ARC-Fi} consistently outperforms all baseline methods, achieving the highest average accuracy of 87.36\%. This confirms that our physics-informed ARC module is genuinely hardware-agnostic. Because ARC exploits the universal spatial diversity of multi-antenna systems rather than memorizing dataset-specific noise patterns, it seamlessly generalizes to the Atheros architecture without requiring any target-domain adaptation.

Furthermore, this evaluation exposes a catastrophic vulnerability in existing vision-inspired paradigms. Baselines such as WiGrunt and WiOpen, which convert CSI data into image-like representations (indicated as \textit{POR+Vis} in Table~\ref{tab:dgcsida}), suffer severe performance collapse, yielding average accuracies of only 25.86\% and 22.99\%, respectively. By destroying the underlying RF phase and amplitude semantics in pursuit of 2D visual patterns, these models fall victim to extreme shortcut learning—failing entirely when the hardware configuration and environment shift. In stark contrast, even the foundational ERM baseline, which preserves signal integrity via standard \textit{POR+LCF}, retains a stable 84.22\% accuracy. This striking dichotomy powerfully validates our core thesis: strictly preserving and exploiting the intrinsic physical properties of RF signals is a non-negotiable prerequisite for scalable, hardware-independent Wi-Fi sensing.

\begin{table}[tbh]
    \caption{
    Cross-hardware Domain Generalization (DG) accuracy (\%) on the CSIDA dataset (Atheros NICs).
    Results are aggregated (Mean$\pm$Std) across all target sub-domains.
    The specific pre-processing pipeline tailored to each method's underlying paradigm is indicated directly below its name.
    The best performance is highlighted in bold.
    }
    \label{tab:dgcsida}
    \centering
    \renewcommand{\arraystretch}{1.2}
    \resizebox{\columnwidth}{!}{
    \begin{tabular}{lccccc}
    \toprule
    \multirow{2}{*}{\textbf{Domain Shift}} & \textbf{ERM} & \textbf{WiSR} & \textbf{WiGrunt} & \textbf{WiOpen} & \textbf{ARC-Fi} \\
    \cmidrule(lr){2-6}
    & \textit{\footnotesize POR+LCF} & \textit{\footnotesize POR+LCF} & \textit{\footnotesize POR+Vis} & \textit{\footnotesize POR+Vis} & \textit{\footnotesize POR+LCF} \\
    \midrule
    Room     & 82.79$\pm$2.55 & 83.49$\pm$3.13 & 25.39$\pm$4.14 & 22.78$\pm$2.89 & \textbf{87.46$\pm$3.42} \\
    Location & 85.68$\pm$2.07 & 85.52$\pm$4.31 & 25.92$\pm$5.44 & 22.81$\pm$3.97 & \textbf{88.07$\pm$3.21} \\
    User     & 84.19$\pm$4.98 & 83.89$\pm$4.14 & 26.26$\pm$4.33 & 23.38$\pm$3.42 & \textbf{86.54$\pm$6.13} \\
    \midrule
    \textbf{Average}& 84.22 & 84.30 & 25.86 & 22.99 & \textbf{87.36} \\
    \bottomrule
    \end{tabular}
    }
\end{table}

\subsection{Ablation Study}
\subsubsection{Impact of Static Path}
\label{sec:ablation}
To evaluate whether static path will influence the performance of domain shift, we conducted experiments comparing model performance with and without static-path removal on Widar cross-domain tasks. This evaluation includes three baseline methods alongside our proposed approach, with results presented in Fig.~\ref{fig:ablation_tdcsi}. 
As discussed in Section~\ref{sec:domainshift}, the results confirm that mitigating environmental confounding is crucial for effectively addressing the cross-domain challenges of CSI data.
Moreover, our method appears to be the most sensitive to static paths, likely because the contrastive loss emphasizes shared components between augmented views, which may inadvertently lead the model to focus on static paths.

\begin{table*}[!tbh]
    \caption{    Ablation study on the Widar dataset, quantifying the contribution of each key component of \textbf{ARC-Fi}: Cross-Domain Antenna Style Transfer (CDAST) and Natural Antenna Response Consistency (NARC). We report the DG accuracy (\%) on nine target domains and the average across them. The results demonstrate that each component is essential for achieving the full model's state-of-the-art performance.
}
    \centering
    \footnotesize
    \setlength{\tabcolsep}{4pt} % reduce horizontal 
    \renewcommand{\arraystretch}{1.4}  
    \begin{tabular}{cllccccccc}
    \toprule
    Model&\multicolumn{2}{c}{Components}&\multicolumn{4}{c}{Widar}&\multicolumn{3}{c}{CSIDA}\\
    \cmidrule(lr){4-7}
    \cmidrule(lr){8-10}
         &CDAST&NARC&Cross Room&Cross Loc&Cross User&Cross Ori&Cross Room&Cross Loc&Cross User\\
    \midrule
         \textbf{ARC-Fi} (Ours)&\cmark&\cmark&\textbf{88.22}&  \textbf{94.08}& \textbf{90.52}&  \textbf{81.12}&  \textbf{87.46}& \textbf{88.07}&  \textbf{86.54}\\
         \textbf{ARC-Fi} w/o CDAST&&\cmark&84.77&  90.92& 87.81&  79.78&  82.65& 86.77&  85.72\\
         \textbf{ARC-Fi} w/o CDAST,NARC&&&76.90&  86.52& 83.63&  70.12&  68.18& 65.46&  61.65\\
         \bottomrule
    \end{tabular}
    \label{tab:ablation_arc}
\end{table*}
% TODO 加一个小节用来表示 表现和超参数alpha的关系
\subsubsection{Effect of Natural ARC and Cross-Domain Antenna Style Transfer}
To validate our design, we performed a thorough ablation study on the Widar dataset, systematically deconstructing \textbf{ARC-Fi} to quantify the contribution of its two key components: Cross-Domain Antenna Style Transfer (CDAST) and Natural Antenna Response Consistency (NARC). The results in Table~\ref{tab:ablation_arc} offer a clear narrative: each component is essential, and they work in synergy.

At the core of \textbf{ARC-Fi}'s success is our NARC module. Removing it alone accounts for the vast majority of the performance gain (5.49\% of the 7.48\% total), confirming our hypothesis that enforcing physics-based consistency is the most critical factor for learning domain-invariant representations.

The complementary roles of CDAST is also evident. Deactivating CDAST, our input-level augmentation, individually reduces average accuracy by 1.99\% This demonstrates a powerful synergy, where CDAST creates more robust features for NARC to regularize.

In conclusion, this study confirms that \textbf{ARC-Fi} is a well-engineered system where NARC provides the foundational generalization capability, which is then significantly enhanced by the synergistic interplay of CDAST.

\subsection{Feature Visualization}

\begin{figure}[!htbp]
    \centering
    \subfloat[WiSR Ratio=1.0]{%
        \includegraphics[width=0.242\textwidth]{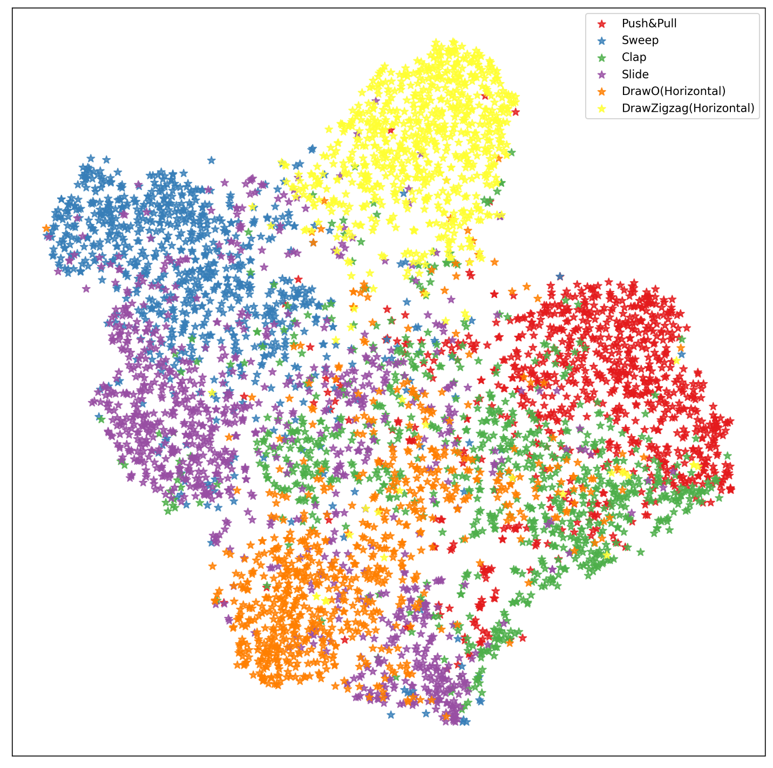}%
        \label{fig:WiSRtsne}%
    }
    \hfill
    \subfloat[WiGrunt Ratio=1.0]{%
        \includegraphics[width=0.242\textwidth]{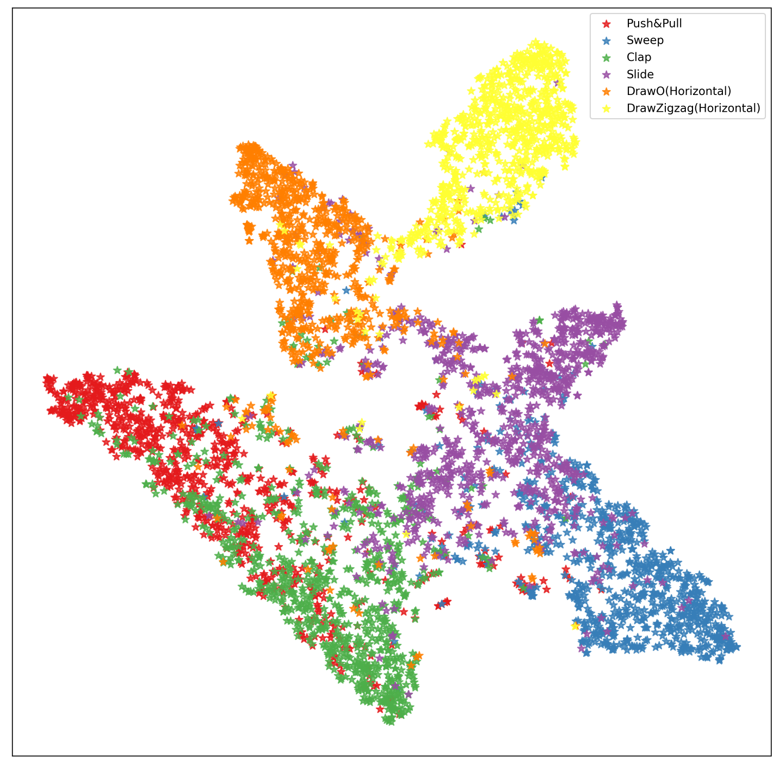}%
        \label{fig:WiGRUNTtsne}%
    }
    \hfill
    \subfloat[Ours Ratio=0.3]{%
        \includegraphics[width=0.242\textwidth]{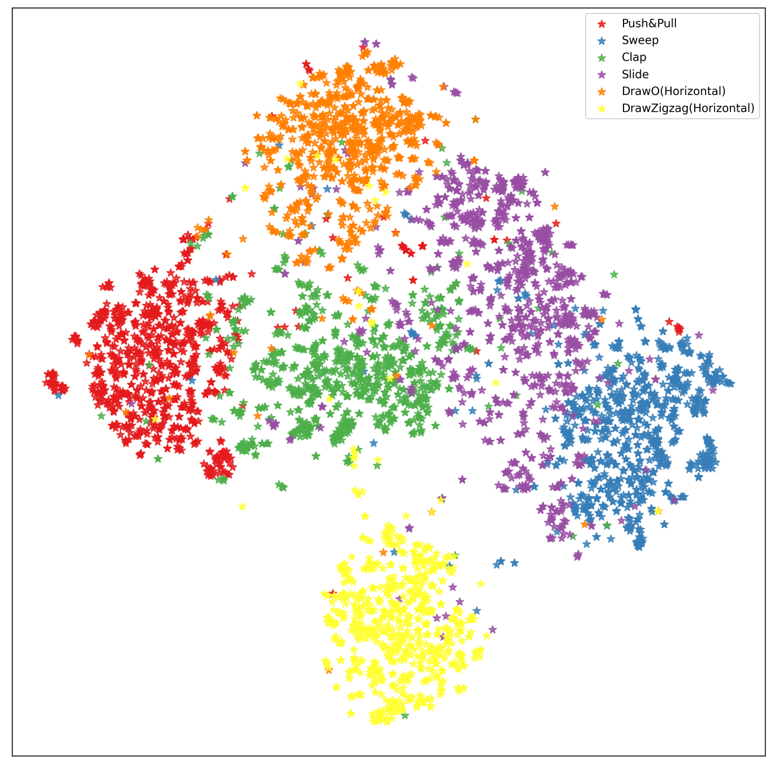}%
        \label{fig:oursr03}%
    }
    \hfill
    \subfloat[Ours Ratio=1.0]{%
        \includegraphics[width=0.242\textwidth]{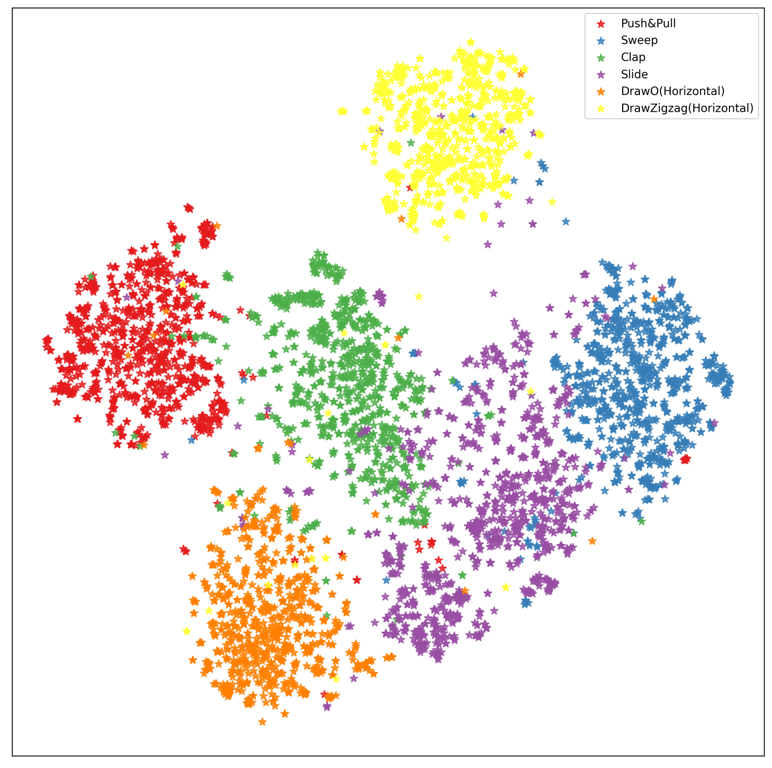}%
        \label{fig:oursr1}%
    }
    \caption{t-SNE visualization of target-domain feature embeddings on Widar. \textit{Ratio} indicates the proportion of labeled source data used during training. Distinct colors represent different gesture categories.}
    \label{fig:tsne}
\end{figure}

To intuitively demonstrate the discriminative power of \textbf{ARC-Fi}, we visualize the target-domain feature embeddings using t-SNE in Fig.~\ref{fig:tsne}. Compared to baselines like WiSR and WiGrunt, \textbf{ARC-Fi} produces notably more compact and well-separated clusters. Remarkably, even when trained with only 30\% of labeled source data (\textit{Ratio = 0.3}), our model yields class boundaries that are visibly clearer and exhibit less inter-class overlap than those of competing approaches trained under fully supervised conditions (\textit{Ratio = 1.0}). This compelling visual evidence further confirms that our framework effectively aligns domain-invariant features and preserves semantic consistency, even under severe label scarcity.

\section{Conclusion}
In this paper, we addressed the critical and intertwined challenges of domain shift and label scarcity in Wi-Fi-based gesture recognition. We proposed \textbf{ARC-Fi}, the first unified framework capable of seamlessly operating across DG and SSDG settings while achieving state-of-the-art performance in UDA scenarios. 

The success of \textbf{ARC-Fi} stems from two tightly coupled innovations: ARC module, which leverages inherent multi-antenna spatial diversity to generate semantically robust training views, and a tailored Semi-Supervised Contrastive Objective that aligns domain-invariant features using both scarce labels and abundant unlabeled data. Extensive evaluations on public datasets demonstrate that \textbf{ARC-Fi} establishes a new benchmark for cross-domain wireless sensing. Notably, our framework operating on a single commodity antenna pair achieves robustness that rivals prior systems relying on complex multi-receiver arrays. 

Crucially, our findings confirm a major conceptual shift: grounding contrastive representation learning in the physical principles of the RF sensing modality—rather than relying on generic, heuristic-based augmentations—effectively prevents paradigm-specific shortcut learning and yields genuinely generalizable representations. 

Despite these advancements, our current framework relies on MIMO systems to generate physical views, precluding its direct application to single-antenna (SISO) devices. Future research will explore constructing physically grounded augmentations from other invariances, such as frequency diversity across subcarriers in SISO setups. Additionally, we plan to extend the \textbf{ARC-Fi} principles beyond discrete gestures to continuous, long-term human activities, such as gait analysis and multi-person interaction, further lowering the barrier for deploying real-world, human-centric sensing systems.

% \clearpage
%\begin{thebibliography}{1}
\bibliographystyle{IEEEtran}  % use IEEE Trans ref style
\bibliography{IEEEabrv,ref}

%{\appendices
%\section*{Proof of the First Zonklar Equation}
%Appendix one text goes here.
% You can choose not to have a title for an appendix if you want by leaving the argument blank
%\section*{Proof of the Second Zonklar Equation}
%Appendix two text goes here.}

\newpage

\vfill

\end{document}